\pdfoutput=1
\documentclass[letterpaper]{article} 
\usepackage{aaai24}  
\usepackage{times}  
\usepackage{helvet}  
\usepackage{courier}  
\usepackage[hyphens]{url}  
\usepackage{graphicx} 
\usepackage{natbib}  
\usepackage{caption} 
\usepackage{algorithm}
\usepackage{algorithmic}

\usepackage{newfloat}
\usepackage{listings}
\DeclareCaptionStyle{ruled}{labelfont=normalfont,labelsep=colon,strut=off} 
\floatstyle{ruled}
\newfloat{listing}{tb}{lst}{}
\floatname{listing}{Listing}
\usepackage{subcaption}
\usepackage{amssymb}
\usepackage{amsmath}
\usepackage{booktabs}
\usepackage{ragged2e} 
\usepackage{booktabs,makecell, multirow, tabularx}

\nocopyright 

\title{Generating Unbiased Pseudo-labels via a Theoretically Guaranteed Chebyshev Constraint to Unify Semi-supervised Classification and Regression}
\author {
    Jiaqi Wu\textsuperscript{\rm 1},
    Junbiao Pang\textsuperscript{\rm 1},
    Qingming Huang\textsuperscript{\rm 2}
}
\affiliations {
    \textsuperscript{\rm 1}Beijing University of Technology\\
    \textsuperscript{\rm 2}University of Chinese Academy of Sciences\\
    qi2019kb@163.com, Junbiao\_pang@bjut.edu.cn, qmhuang@ucas.ac.cn
}

\usepackage{bibentry}

\begin{document}

\maketitle

\begin{abstract}
Both semi-supervised classification and regression are practically challenging tasks for computer vision. However, semi-supervised classification methods are barely applied to regression tasks. Because the threshold-to-pseudo label process (T2L) in classification uses confidence to determine the quality of label. It is successful for classification tasks but inefficient for regression tasks. In nature, regression also requires unbiased methods to generate high-quality labels. On the other hand, T2L for classification often fails if the confidence is generated by a biased method. To address this issue, in this paper, we propose a theoretically guaranteed constraint for generating unbiased labels based on Chebyshev's inequality, combining multiple predictions to generate superior quality labels from several inferior ones. In terms of high-quality labels, the unbiased method naturally avoids the drawback of T2L. Specially, we propose an Unbiased Pseudo-labels network (UBPL network) with multiple branches to combine multiple predictions as pseudo-labels, where a Feature Decorrelation loss (FD loss) is proposed based on Chebyshev constraint. In principle, our method can be used for both classification and regression and can be easily extended to any semi-supervised framework, e.g. Mean Teacher, FixMatch, DualPose. Our approach achieves superior performance over SOTAs on the pose estimation datasets Mouse, FLIC and LSP, as well as the classification datasets CIFAR10/100 and SVHN.
\end{abstract}

\section{Introduction}

Classification and regression are two important tasks in computer vision that have achieved great success in numerous real-world applications \cite{krizhevsky2009learning, he2016deep, sun2019deep, wang2013approach, wang2016mining}, especially in supervised learning scenarios that utilize large amounts of high-quality labeled data. However, labeled data is difficult to obtain in many domains, and manually labeling data is a time-consuming and costly task. For example, in medical tasks, multiple human experts need to perform time-consuming analysis to label samples. Compared to labeled data, unlabeled data is more accessible and less costly. Therefore, semi-supervised learning (SSL), which can leverage a large amount of unlabeled data to improve the performance of learning with a limited amount of labeled data, has attracted a lot of attention.

\begin{figure}[t]
\centering
\begin{subfigure}{0.49\columnwidth}
	\centering
	\includegraphics[width=1.0\linewidth]{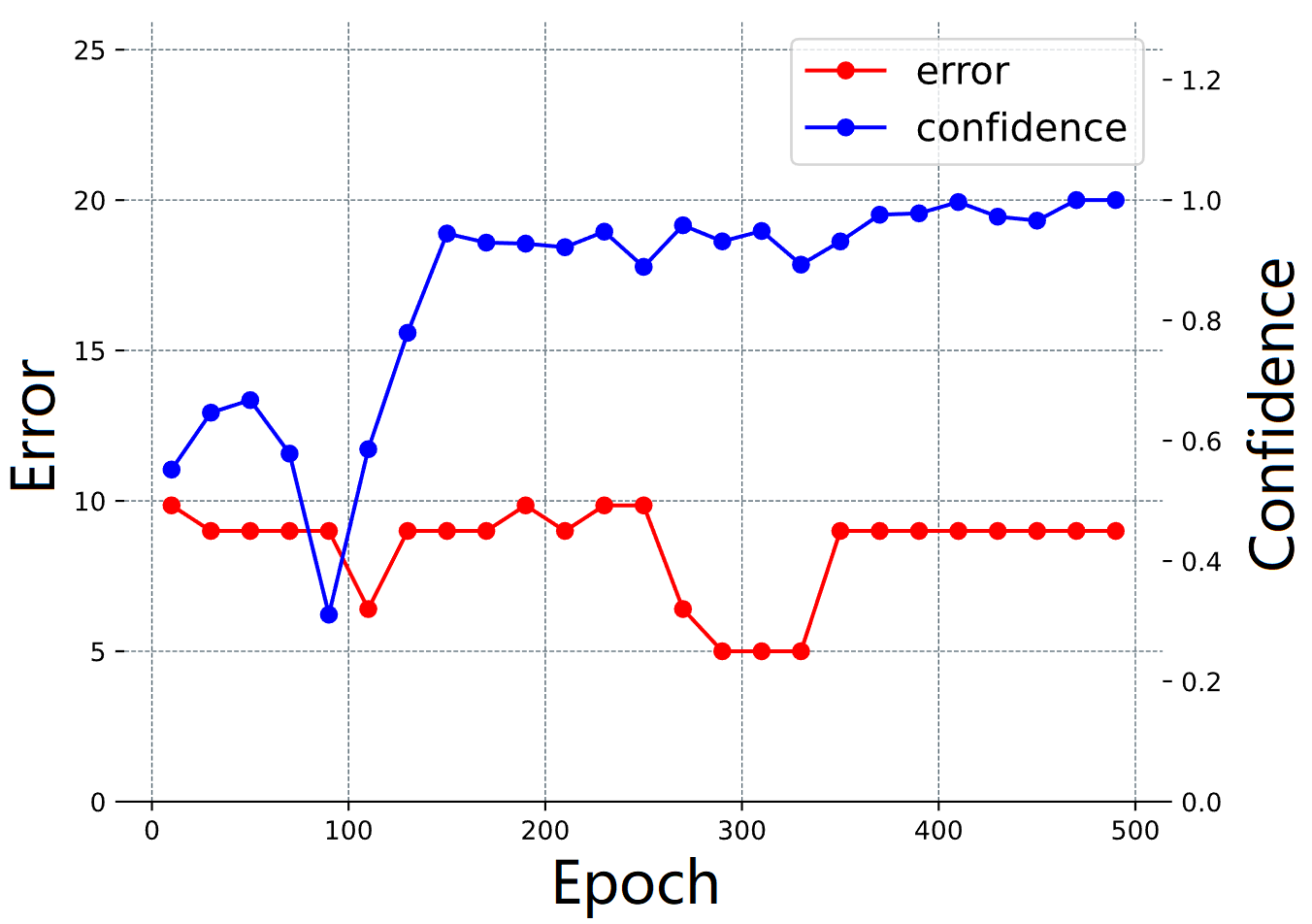}
\end{subfigure}
\hfill
\begin{subfigure}{0.49\columnwidth}
	\centering
	\includegraphics[width=1.0\linewidth]{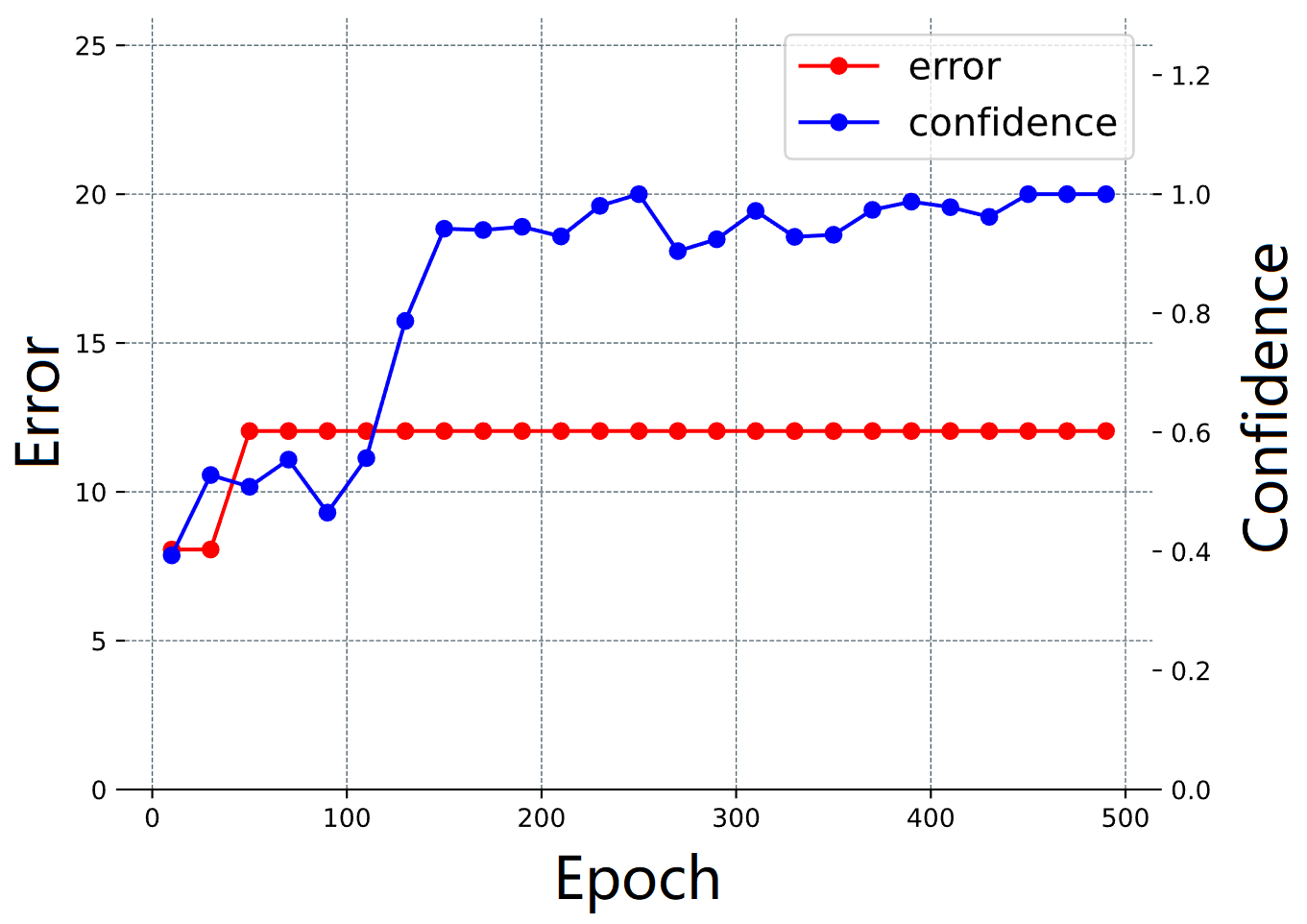}
\end{subfigure}
\caption{Error and confidence curves of two pseudo-labels generated by Mean Teacher \cite{tarvainen2017mean}, a typical semi-supervised classification method, in the pose estimation task.}
\label{fig1_BiasedMethod}
\end{figure}

The Pseudo-labeling (PL) method is an essential type of semi-supervised classification methods. The PL methods \cite{xie2020self, sohn2020fixmatch, zhang2021flexmatch, wang2022freematch} is to use the pseudo-labels generated by the model to train itself during training. Thus the PL method relies on the confidence of the pseudo-label, and the pseudo-label with strong confidence score is selected out and added to the training data set through the threshold-to-pseudo label process (T2L). In the classification task, the T2L is justified. This is because confidence, which is the probability of a model's prediction, can be viewed as how confident the model is about the outcome of the prediction, which can effectively assess the quality of the prediction. 

Nonetheless, semi-supervised classification methods are barely applied to regression tasks, since the original T2L in classification is not suitable for regression tasks. Take pose estimation as an example. In pose estimation tasks, the dominant approach is to use heatmaps to predict keypoints. Compared to classification, the confidence in the heatmap represents only the possibility that a location of the heatmap is a keypoint, not the quality of that prediction. As shown in Fig. \ref{fig1_BiasedMethod}, the error of the generated pseudo-label does not decrease when the confidence of the pseudo-label increases. This indicates that the SSL method is not unbiased in the regression task, and the generated confidence scores do not effectively assess the quality of the pseudo-labels. Therefore, for the regression task, semi-supervised classification methods fail to generate high-quality pseudo-labels.

The "non-unbiased" problem also exists in classification. Currently, semi-supervised classification methods focus more on how to improve the dynamic control strategy for confidence thresholds, hoping to select out high-quality pseudo-labels through more reasonable dynamic thresholds. However, these methods all ignore the point of whether the confidence itself is plausible, after all, the confidence is also predicted by the method. If the method itself is biased, the confidence of the generated predictions may be inaccurate, resulting in inaccurate pseudo-labels selected by T2L. Therefore, "non-unbiased" can not be ignored in the semi-supervised classification task.

To solve the "non-unbiased" problem, we propose a theoretically guaranteed constraint for generating unbiased labels based on Chebyshev's inequality, encouraging combining multiple predictions to generate superior quality labels from several inferior ones. Specially, we propose an Unbiased Pseudo-labels network (UBPL network) with multiple branches to combine multiple predictions as pseudo-labels, where a Feature Decorrelation loss (FD loss) is proposed based on Chebyshev constraint. In principle, our method can be used for both classification and regression and can be easily extended to any semi-supervised framework, e.g. Mean Teacher \cite{tarvainen2017mean}, FixMatch \cite{sohn2020fixmatch}, DualPose \cite{xie2021empirical}. Our approach achieves superior performance over SOTAs on the pose estimation datasets Mouse, FLIC \cite{sapp2013modec} and LSP \cite{johnson2011learning}, as well as the classification datasets CIFAR10/100 \cite{krizhevsky2009learning} and SVHN \cite{netzer2011reading}.

The regression project code and Mouse dataset are publicly available on \url{https://github.com/Qi2019KB/UBPL-PoseEstimation}. And, the classification project code is publicly available on \url{https://github.com/Qi2019KB/UBPL-Classification}

The key \textbf{contributions} of this work are: 

\begin{itemize}
	\item We propose a theoretically guaranteed constraint for generating unbiased labels based on Chebyshev's inequality, which encourages combining multiple predictions to generate superior quality labels from several inferior ones. Benefiting from this, semi-supervised classification methods can be well applied to regression tasks, which implies unifying semi-supervised classification and regression.
	\item We propose a Feature Decorrelation loss (FD loss) based on the Chebyshev constraint to maximize the diversity. The FD loss effectively improves the accuracy and sustainability of the generated pseudo-labels.
\end{itemize}

\section{Related Work}

\subsection{Semi-supervised Classification} 
Semi-supervised learning (SSL) can achieve better training results by using a limited amount of labeled data and a large amount of unlabeled data. Because it can effectively reduce the need for labeled data, SSL is well suited for real-world application scenarios. There have been a number of excellent results in semi-supervised classification.

The first type is the Pseudo-label (PL) method. The PL method trains itself during training using pseudo-labels generated by the model. In general, the PL method uses the threshold-to-pseudo label process (T2L) to select out high-quality pseudo-labels to add to the training set. FixMatch \cite{sohn2020fixmatch} is a landmark that greatly improves the performance of SSL. FixMatch uses the predictions of weakly augmented samples as pseudo-labels to supervise the predictions of strongly augmented samples, thereby obtaining an additional supervised signal from unlabeled samples. FlexMatch \cite{zhang2021flexmatch} proposed the idea of adaptive the threshold based on FixMatch. FreeMatch \cite{wang2022freematch} adaptively adjusts the threshold based on the learning state of the model and uses an adaptive class fairness regularization penalty to encourage the model to make diversified predictions in the early training phase. These methods are mainly improved from the perspective of threshold in T2L, focusing on how to select better pseudo-labels. 

Another Type is the Consistency Regularization (CR) method \cite{sajjadi2016regularization, laine2016temporal, tarvainen2017mean, miyato2018virtual, bachman2014learning, ke2019dual}. The CR method encourages the model to produce approximate predictions for different perturbations in the same sample and minimizes the prediction difference to train the model. Such perturbation includes data enhancement \cite{sajjadi2016regularization}, adversarial sample \cite{miyato2018virtual}, model perturbation \cite{bachman2014learning}, etc. A typical approach to CR methods is the teacher-student structure \cite{tarvainen2017mean}, which uses the predictions of the teacher to supervise the students through consistency constraints. Similar to the PL method, since the target is generated by the model itself, the target is imprecise. The CR method also requires T2L to select high-quality pseudo-labels, whose confidence predicted from teacher is higher than that from student.

Although the above methods all use T2L to generate pseudo-labels, there are still shortcomings. This is because, when confidence thresholds are used for T2L, it is not considered whether the confidence scores generated by the bias method are accurate or sufficiently reflective of the quality of the pseudo-labels.

\subsection{Semi-supervised Regression}

At present, the research on semi-supervised regression method is much less than semi-supervised classification. Since regression methods predict true-valued outputs, whereas classification methods typically predict within a finite number of categories, regression is far more difficult than classification. However, this difficulty is more prominent in semi-supervised regression tasks, which makes the confidence and the corresponding prediction result more imprecise. As a result, the pseudo-labels selected with T2L are also less accurate.

Take pose estimation as an example. DualPose \cite{xie2021empirical} is a milestone in the field of semi-supervised pose estimation. DualPose addresses the problem of model collapsing in 2D human pose estimation by the consistent regularization loss and proposes to solve the problem by constructing pairs of samples with different prediction difficulties.  However, the confidence of a heatmap only indicates the possibility that a certain location in the heatmap is a keypoint, not the quality of the prediction. As a consequence, if a method is biased, the pseudo-label selected by T2L cannot provide an additional effective supervision signal due to the inaccurate generated pseudo-label.

To further confirm this, we use Mean Teacher for semi-supervised pose estimation, and the confidence and error of the pseudo-labels generated by Teacher are shown in Fig. \ref{fig1_BiasedMethod}. As the confidence of the pseudo-label increases, the error of the pseudo-label does not decrease. These results indicate that the SSL method is not unbiased in the regression task, and the generated confidence scores are not effective in evaluating the quality of the pseudo-labels. Therefore, regression also requires unbiased methods to generate high-quality labels.

\subsection{Multi-branch structure in SSL}
Several works have applied multi-branch structure to SSL. \cite{ke2019dual} finds that, under the influence of EMA mechanism, the model parameters of Mean Teacher and student become more and more similar, which leads to the convergence of their predictions and the loss of the supervision effectiveness of consistency constraints. To this end, in \cite{ke2019dual}, an additional student is added to reduce this impact. Also considering that the model parameters of teacher and student tend to be consistent, \cite{tang2021humble} proposes a two-teacher structure, and EMA is used to update the parameters of two teachers alternately, so as to increase the difference between the two teachers' parameters. However, both approaches have drawbacks. Although the additional student in \cite{ke2019dual} reduces the risk of parameter consistent, compared with teacher, student has lower prediction accuracy and stability, and is more likely to introduce noise, reducing the accuracy of predictions. In \cite{tang2021humble}, although the diversity has been improved to some extent, the trend of parameter consistent between teacher and student has not changed substantially. 

Considering these, we propose a network with multiple branches to combine multiple predictions as pseudo-labels, where a Feature Decorrelation loss (FD loss) is proposed based on Chebyshev constraint.

\section{Method}

\subsection{Theoretically Guaranteed Chebyshev Constraint}
To solve the "non-unbiased" problem, we defined a theoretically guaranteed chebyshev constraint.

Suppose we use $ T $ predictors $ H = \{h_1, h_2, \cdots, h_t\} $ to build the ensemble architecture. The prediction of each predictor is $ h_1(x),\cdots,h_t(x) $, as $ h_1,\cdots,h_t $, and the ensemble prediction is $ \overline{H}(x) $, as $ \overline{H} $. The aim of ensemble learning is to make the ensemble prediction $ \overline{H} $ as close as possible to the optimal prediction $ H^{*}(x) $, as $ H^{*} $.

One of the intuitions of ensemble learning is that when the predictors are trained synchronously using the same data set, they can be considered to be close to the optimal prediction $ H^{*} $ at a similar optimization rate. Therefore, we assume that:

\begin{equation}\label{eqt10_Ensemble1}
	h_1 - H^{*} \approx \cdots, h_2 - H^{*} \approx \overline{H} - H^{*},
\end{equation}

Based on \eqref{eqt10_Ensemble1}, we can easily get:

\begin{equation}\label{eqt11_Ensemble2}
	\mathbb{E}(H^{*}) = \frac{1}{T}\sum_{t=1}^{T}h_t = \overline{H} ,
\end{equation}

According to Chebyshev inequality, we define that the error between ensemble prediction $ \overline{H} $ and optimal prediction $ H^{*} $ can be approximately expressed as:

\begin{equation}\label{eqt12_Ensemble3}
	\begin{aligned}
		err &= P\{|\overline{H} - H^{*}|\} \\
		&= P\{|H^{*} - \mathbb{E}(H^{*})| \ge \varepsilon \} \\
		&\le \frac{1}{\varepsilon^2}\text{var}(H^{*})
	\end{aligned}
\end{equation}

Since $ \text{var}(H^{*}) \le \text{var}(\overline{H}) $, then:

\begin{equation}\label{eqt13_Ensemble4}
	\begin{aligned}
		err &\le \frac{1}{\varepsilon^2}\text{var}(\overline{H}) \\
		&\le \frac{1}{\varepsilon^2}\text{var}\{\frac{\sum_{t=1}^{T}h_t}{T}\} \\
		&\le \frac{1}{\varepsilon^2}\frac{1}{T^2}\left[ \underbrace{\sum_{t=1}^{T}\text{var}(h_t)}_{\text{var}(H)} + \underbrace{\sum_{t=1}^{T}\sum_{t=1, j\neq t}^{T}2\text{covar}(h_t, h_j)}_{\text{covar}(H)} \right] ,
	\end{aligned}
\end{equation}

According to \eqref{eqt13_Ensemble4}, factors affecting the ensemble prediction error include stability of each predictor ($\text{var}(H)$) and diversity of ensemble ($\text{covar}(H)$). 

(a) \textbf{Stability} The variance represents sensitivity of the predictor itself to different augmentation (or perturbation) of the same sample. Small variance implies the small prediction error, which means that each predictor needs to be stable, which is consistent with \eqref{eqt13_Ensemble4}.

(b) \textbf{Diversity} Correlations between predictors model ensemble diversity. As shown in \eqref{eqt13_Ensemble4}, the lower the correlation between the predictors, the smaller the ensemble prediction error, which is consistent with the effect of diversity.

\subsection{Overview}

\begin{figure}[h]
	\centering
	\includegraphics[width=0.9\columnwidth]{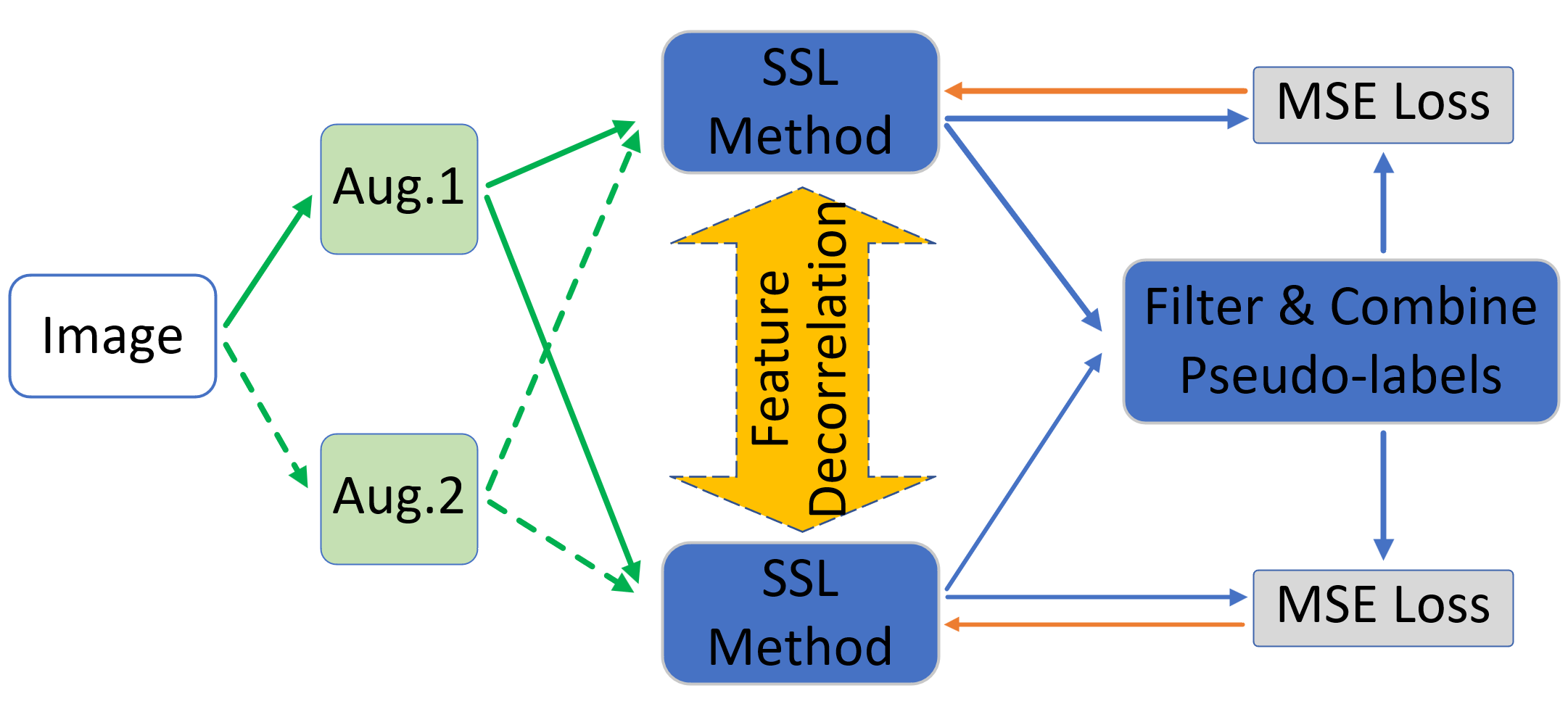}
	\caption{Overview of the proposed Unbiased Pseudo-labels network (UBPL network).}
	\label{fig2_ModelStructure}
\end{figure}

We propose an Unbiased Pseudo-labels network (UBPL network) with multiple branches to combine multiple predictions as pseudo-labels, so as to obtain the better supervised signals from unlabeled data.

The UBPL network, shown in Fig. \ref{fig2_ModelStructure}, consists of two branches with the same structure (only the initialization strategies for the model parameters differ). Each branch is a semi-supervised learning process. This is a general structure, so the SSL method here can be any of the mainstream SSL methods including Mean Teacher, FixMatch, etc. 

In each branch, we input two different data augmentation samples, a hard prediction sample generated by strong data augmentation and an easy prediction sample generated by weak data augmentation. The output predictions of unlabeled samples from the two branches are then filtered and combined, and the final ensemble predictions are used as pseudo-labels to supervise the training of the two branches separately.

In addition, considering the problem of model homogenization, we propose a Feature Decorrelation Loss (FD Loss) based on Chebyshev constraints to maximize diversity and maintain the difference between branches.

\subsection{Feature Decorrelation Loss}

\begin{figure}[h]
	\centering
	\includegraphics[width=0.9\columnwidth]{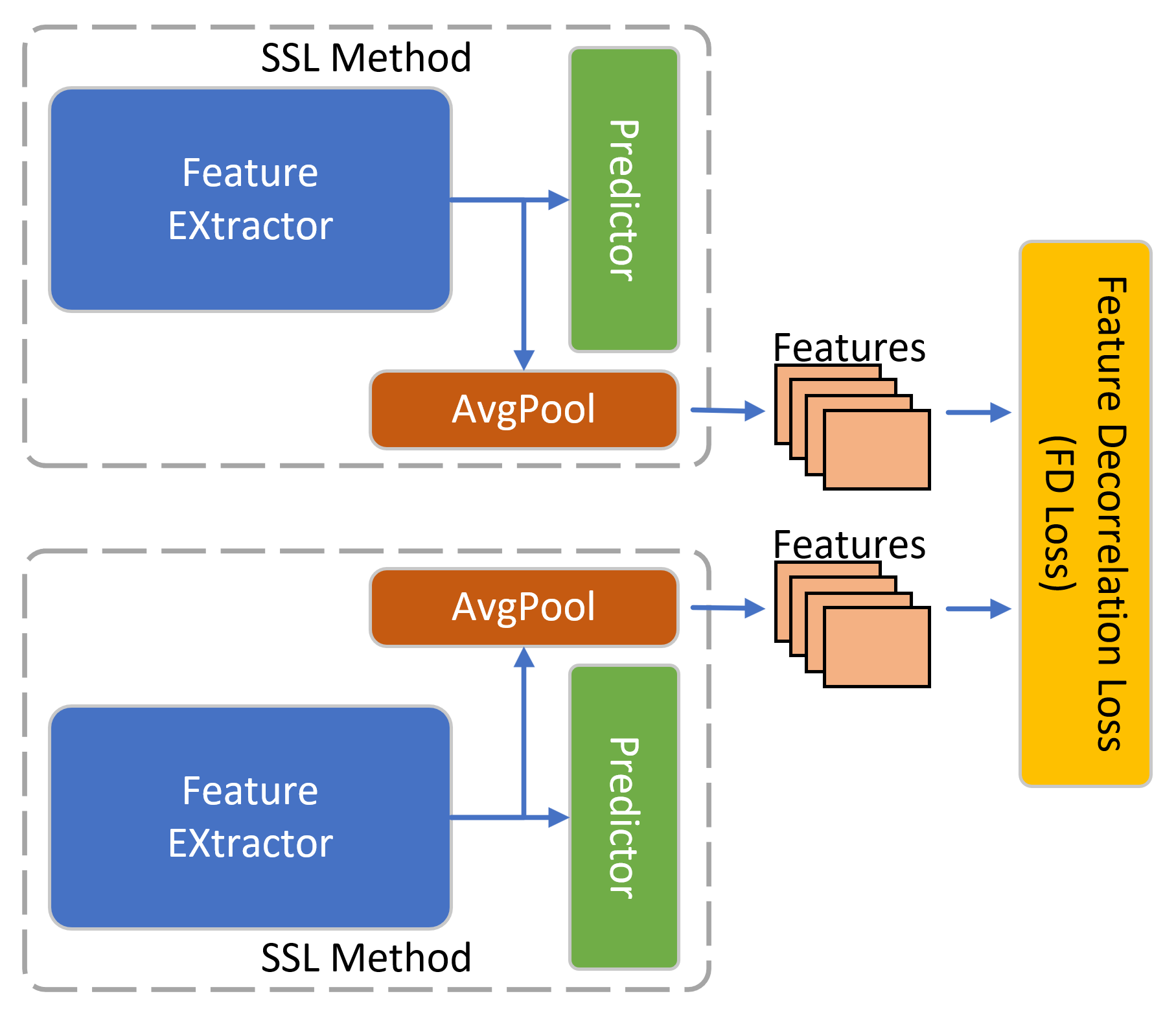}
	\caption{The structure of Feature Discrepancy loss (FD loss). To compute feature correlations between two branches, we output features for classification prediction. In addition, to reduce the amount of computation and improve the computational robustness of feature correlation, we perform Average Pooling on the output features. Then, the FD loss is used to minimize the two corresponding features.}
	\label{fig3_FDL_Structure}
\end{figure}

Based on the above discussion, we propose a Feature Decorrelation loss (FD loss) is proposed based on Chebyshev constraint.

The features extracted by the model can be considered as salient information discovered by the model from the input sample in order to accurately predict this sample. If the two models find different salient information for the same sample, one can argue that the two models have different concerns. The difference in concerns reflects the low correlation between the two models.  It is also reflected in \eqref{eqt13_Ensemble4}.

We use the labeled data to calculate the FD loss.  Moreover we reshape the feature ($ C*H*W $), which is outputted from the AvgPool layer in the model of each branch, to $ C*(H*W) $. $ C $, $ H $, $ W $ are the channel, height, width of feature, respectively.  Thus the FD loss $ L_{fd} $ is as follows:

\begin{equation}\label{eqt13_FDLoss}
	L_{fd} = \frac{1}{N_B}\frac{1}{C}\beta_{fd}\sum_{i=1}^{N_B}\sum_{c=1}^{C}\text{cov}(g_{br_1}\big(\big[\omega(x_i^l)\big]_c\big), g_{br_2}\big(\big[\omega(x_i^l)\big]_c\big))) ,
\end{equation}
where $ g_{br_1} $ and $ g_{br_2} $ are the features outputted from the AvgPool layer of models $ f_{br_1} $ and $ f_{br_2} $ of two branches, respectively. $ \big[\cdot\big]_c $ is the $ c $ channel's value of feature. Due to the small value of the FD loss, we introduce an expansion factor $ \beta_{fd} $ to balance the FD loss with the other loss functions.

\subsection{Total Loss}

So, the total loss of each branch is as follows:

\begin{equation}\label{eqt14_TotalLoss}
	L_{br} = \lambda_{ssl} L_{ssl} + \lambda_{pse} L_{pse} + \lambda_{fd} L_{fd},
\end{equation}
where $ L_{ssl} $ is the loss of the SSL method in the branch, which varies depending on the SSL method used. 

\subsection{SSL Method Selection}
In general, we can decompose the semi-supervised method into two parts. One is a supervised learning process based on labeled data. The other part is the unsupervised learning process based on unlabeled d (some are also called self-supervised learning processes). Consistent regularization methods and pseudo-labeling methods are typical representatives of the unsupervised learning process, and their essence is to obtain additional trusted supervised signals from unlabeled samples to improve the training performance of semi-supervised learning. Thus, we can express the loss function of the semi-supervised method $ L_{ssl} $ uniformly as follows:

\begin{equation}\label{eqt14_TotalLoss}
	L_{ssl} = L_{sup} + L_{unsup},
\end{equation}
where $ L_{sup} $ is the loss of supervised learning process, and $ L_{unsup} $ is the loss of unsupervised learning process.

In a follow-up experiment, we will combine Mean Teacher, FixMatch, DualPose, respectively, and perform in both classification and pose estimation.

\section{Experiments on Pose Estimation}

We perform semi-supervised pose estimation experiments to validate the effectiveness of our method on semi-supervised regression tasks.

\subsection{Implementation Details}

\begin{table*}[t]
	\centering
	\begin{tabular}{@{}lcccc@{}}
		\toprule 
		\multirow{3}{*}{Method} & \multicolumn{4}{c}{Mouse} \\
		& \multicolumn{2}{c}{30/100} & \multicolumn{2}{c}{30/200} \\
		& MSE & PCK@0.2 & MSE & PCK@0.2 \\
		\midrule
		Supervised                            & 9.59 $\pm$ 0.05       & 0.264 $\pm$ 0.005       & 8.45 $\pm$ 0.12       & 0.266 $\pm$ 0.012 \\
		\midrule
		Mean Teacher \cite{tarvainen2017mean} & 4.60 $\pm$ 0.14       & 0.591 $\pm$ 0.007       & 3.92 $\pm$ 0.00       & 0.678 $\pm$ 0.000 \\
		Mean Teacher + UBPL                   & \pmb{4.39} $\pm$ 0.01 & \pmb{0.617} $\pm$ 0.003 & \pmb{3.65} $\pm$ 0.00 & \pmb{0.719} $\pm$ 0.001 \\
		\midrule
		DualPose \cite{xie2021empirical}      & 4.74 $\pm$ 0.00       & 0.575 $\pm$ 0.002       & 3.90 $\pm$ 0.02       & 0.683 $\pm$ 0.002 \\
		DualPose + UBPL                       & \pmb{4.59} $\pm$ 0.02 & \pmb{0.579} $\pm$ 0.004 & \pmb{3.75} $\pm$ 0.07 & \pmb{0.686} $\pm$ 0.001 \\
		\bottomrule
		\toprule 
		\multirow{3}{*}{Method} & \multicolumn{4}{c}{FLIC} \\
		& \multicolumn{2}{c}{30/100} & \multicolumn{2}{c}{30/200} \\
		& MSE & PCK@0.5 & MSE & PCK@0.5 \\
		\midrule
		Supervised                            & 50.70 $\pm$ 1.07       & 0.441 $\pm$ 0.009       & 54.03 $\pm$ 0.06       & 0.396 $\pm$ 0.001 \\
		\midrule
		Mean Teacher \cite{tarvainen2017mean} & 43.14 $\pm$ 0.04       & 0.516 $\pm$ 0.005       & 48.67 $\pm$ 0.31       & 0.449 $\pm$ 0.001 \\
		Mean Teacher + UBPL                   & \pmb{40.29} $\pm$ 0.28 & \pmb{0.529} $\pm$ 0.002 & \pmb{47.03} $\pm$ 0.59 & \pmb{0.450} $\pm$ 0.003 \\
		\midrule
		DualPose \cite{xie2021empirical}      & 41.89 $\pm$ 0.67       & 0.514 $\pm$ 0.009       & 46.83 $\pm$ 0.52       & 0.453 $\pm$ 0.003 \\
		DualPose + UBPL                       & \pmb{41.51} $\pm$ 0.08 & \pmb{0.520} $\pm$ 0.003 & \pmb{46.18} $\pm$ 0.56 & \pmb{0.454} $\pm$ 0.001 \\
		\bottomrule
		\toprule 
		\multirow{3}{*}{Method} & \multicolumn{4}{c}{LSP}\\
		& \multicolumn{2}{c}{100/500} & \multicolumn{2}{c}{200/500} \\
		& MSE & PCK@0.5 & MSE & PCK@0.5 \\
		\midrule
		Supervised                            & 52.22 $\pm$ 0.81       & 0.102 $\pm$ 0.001       & 47.02 $\pm$ 0.71       & 0.150 $\pm$ 0.004 \\
		\midrule
		Mean Teacher \cite{tarvainen2017mean} & 49.56 $\pm$ 0.03       & \pmb{0.146} $\pm$ 0.000 & 41.61 $\pm$ 0.54       & \pmb{0.220} $\pm$ 0.001 \\
		Mean Teacher + UBPL                   & \pmb{45.48} $\pm$ 0.27 & 0.130 $\pm$ 0.003       & \pmb{39.60} $\pm$ 0.01 & 0.189 $\pm$ 0.001 \\
		\midrule
		DualPose \cite{xie2021empirical}      & 46.35 $\pm$ 0.46       & \pmb{0.131} $\pm$ 0.000 & \pmb{39.81} $\pm$ 0.10 & \pmb{0.183} $\pm$ 0.001 \\
		DualPose + UBPL                       & \pmb{45.73} $\pm$ 0.62 & 0.129 $\pm$ 0.002       & 39.82 $\pm$ 0.07       & 0.182 $\pm$ 0.003 \\
		\bottomrule
	\end{tabular}
	\caption{The MSE and PCK of semi-supervised pose estimation on Mouse, FLIC and LSP datasets. 'Supervised' denotes training a single Pose Model (Stacked Hourglass \cite{newell2016stacked}) in supervised manner with only labeled data. '+ UBPL' denotes using our UBPL network. '30/100' denotes the data count is 100 with 30 labeled data.}
	\label{tab1_PoseMainResults}
\end{table*} 

\subsubsection{Dataset}

Our experiments are mainly performed on the following three datasets: FLIC \cite{sapp2013modec}, LSP \cite{johnson2011learning} dataset, and our own collected Mouse dataset. FLIC and LSP datasets are open-source human pose datasets commonly used in human pose estimation experiments. Moreover, to present our approach from more perspectives, we introduce the Mouse dataset. It is experimental data on the sniffing movements of laboratory mice captured in a realistic experimental environment.

\subsubsection{Details}

\begin{figure}[t]
	\centering
	\begin{subfigure}{0.49\columnwidth}
		\centering
		\includegraphics[width=1.0\linewidth]{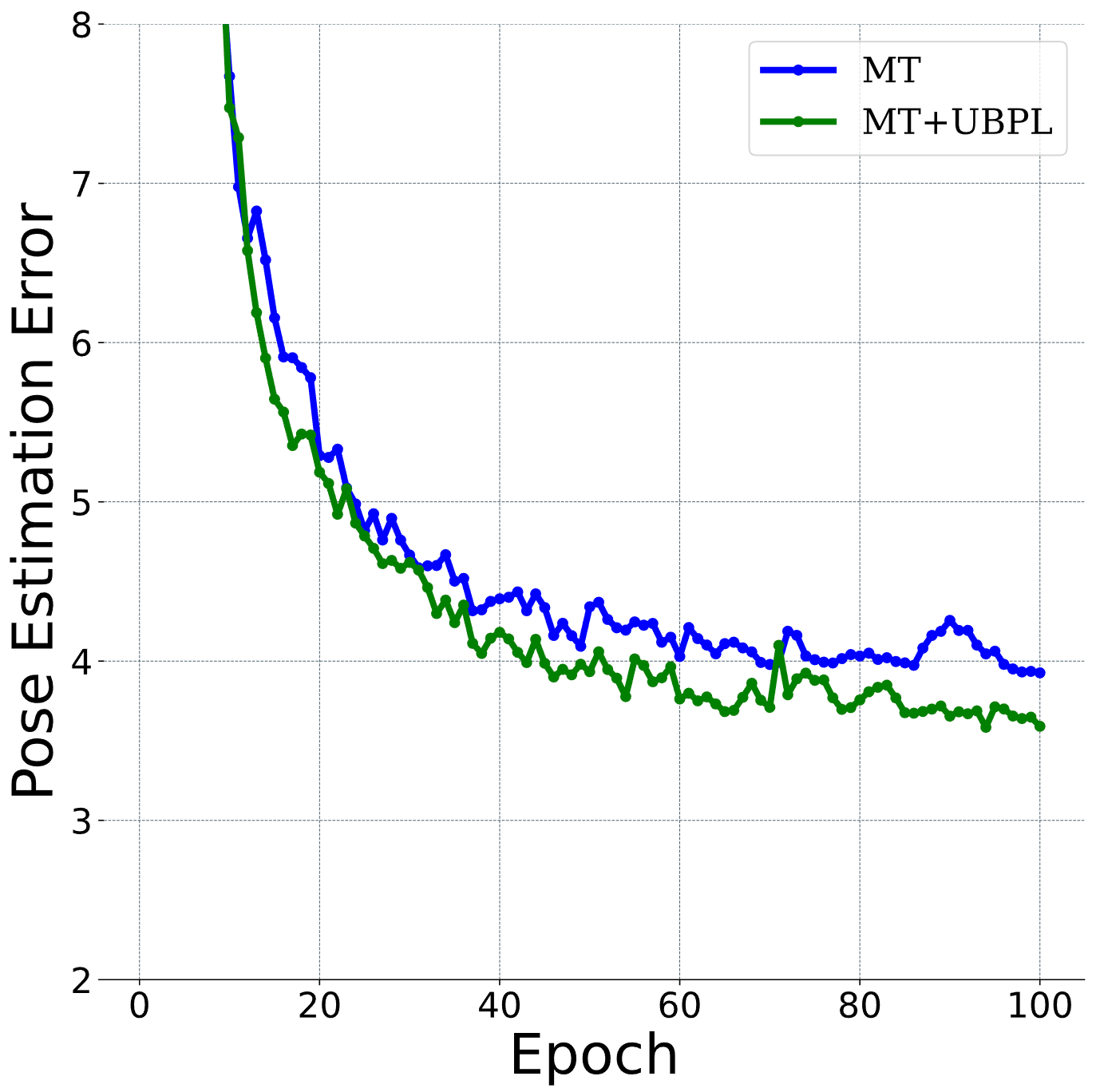}
	\end{subfigure}
	\hfill
	\begin{subfigure}{0.49\columnwidth}
		\centering
		\includegraphics[width=1.0\linewidth]{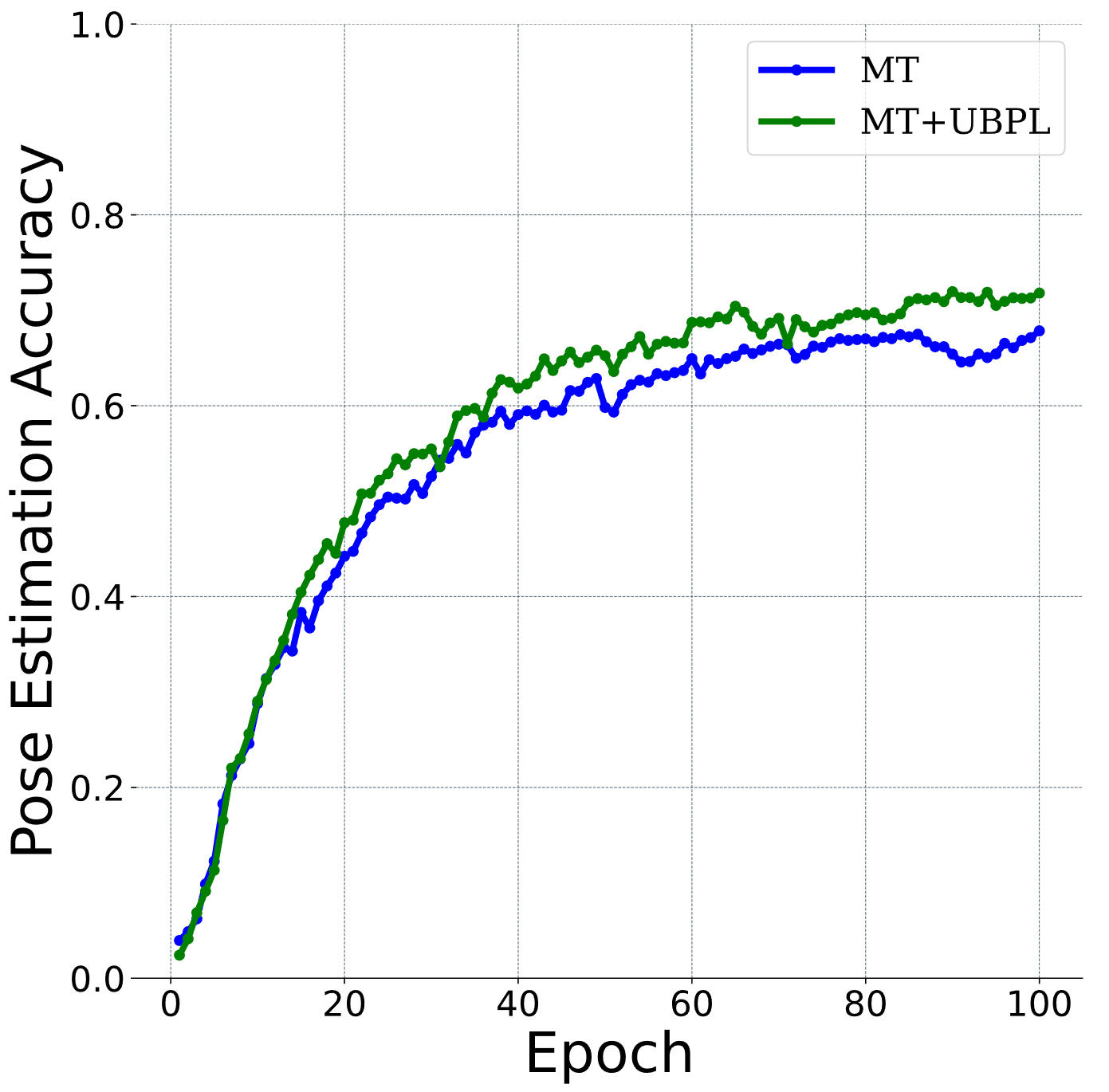}
	\end{subfigure}
	\caption{Error and accuracy curves for 'MT' and 'MT+UBPL' on the Mouse (30/200).}
	\label{fig4_PoseResults}
\end{figure}

We conduct experiments on the Mouse, FLIC and LSP dataset and extensively study the training performance under various labeled data amounts. In all experiments, the pose model is Stacked Hourglass \cite{newell2016stacked}.

For fair comparison, the same hyper-parameters are used in all experiments, except that the data augmentation varies depending on the semi-supervised method used. Specifically, all experiments are performed with an Adam with weight decay 0.The learning rate is 0.00025. The batch size is 4 with 2 labeled data. The confidence threshold $ \tau $ is 0.95. The decay value for the Exponential Moving Average (EMA) \cite{tarvainen2017mean} is 0.999.

To verify the universality of UBPL, we conduct semi-supervised pose estimation experiments using both Mean Teacher and DualPose as SSL methods. 

In terms of data augmentation, Mean Teacher generates two augmented sample pairs with similar prediction difficulty. The data enhancement methods used include: random rotation ($+/- 30$ degrees), random scaling ($ 0.75 - 1.25 $) and random horizontal flip.

And, DualPose uses two data enhancement strategies of varying strength to generate a set of sample pairs with varying prediction difficulty. The strong data augmentation include: random rotation ($+/- 30$ degrees), random scaling ($ 0.75 - 1.25 $) and random horizontal flip. And the weak data augmentation include: random rotation ($+/- 5$ degrees), random scaling ($ 0.95 - 1.05 $) and random horizontal flip.

$ \lambda_{ssl} $ is 10, $ \lambda_{pse} $ is 10. $ L_{fd} $ is 1.

Due to hardware limitations, all experiments were trained for 100 epochs. We used a fixed random seed (1388) to obtain accurate and reliable experimental results.

\subsection{Comparison with Existing Methods}

The MSE and PCK of our method on Mouse, FLIC, LSP datasets are shown in Table. \ref{tab1_PoseMainResults}. The results show that our approach achieves better results on most of the baseline datasets. This is especially the case when the total sample count is small. These results show that our approach improves the performance of existing semi-supervised methods by making towards unbiased predictions to provide more accurate and informative pseudo-labels.

As shown in Fig. \ref{fig4_PoseResults}, our UBPL performs better throughout the training period in the pose estimation task, indicating that UBPL is towards unbiased and can generate better pseudo-labels.

\subsection{Ablation Studies}

\begin{table}[h]
	\centering
	\begin{tabular}{@{}lcc@{}}
		\toprule 
		\multirow{2}{*}{Method} & \multicolumn{2}{c}{Mouse (30/100)} \\
		& MSE & PCK@0.2 \\
		\midrule
		Supervised                            & 9.59 $\pm$ 0.05       & 0.264 $\pm$ 0.005 \\
		\midrule
		Mean Teacher                          & 4.60 $\pm$ 0.14       & 0.591 $\pm$ 0.007 \\
		Mean Teacher + UBPL (noFDL)           & 4.55 $\pm$ 0.03       & 0.608 $\pm$ 0.000 \\
		Mean Teacher + UBPL                   & \pmb{4.39} $\pm$ 0.01 & \pmb{0.617} $\pm$ 0.003 \\
		\midrule
		DualPose                              & 4.74 $\pm$ 0.00       & 0.575 $\pm$ 0.002 \\
		DualPose + UBPL (noFDL)               & 4.63 $\pm$ 0.03       & 0.576 $\pm$ 0.003 \\
		DualPose + UBPL                       & \pmb{4.59} $\pm$ 0.02 & \pmb{0.579} $\pm$ 0.004 \\
		\bottomrule
		\toprule 
		\multirow{2}{*}{Method} & \multicolumn{2}{c}{Mouse (30/200)} \\
		& MSE & PCK@0.2 \\
		\midrule
		Supervised                            & 8.45 $\pm$ 0.12       & 0.266 $\pm$ 0.012 \\
		\midrule
		Mean Teacher                          & 3.92 $\pm$ 0.00       & 0.678 $\pm$ 0.000 \\
		Mean Teacher + UBPL (noFDL)           & 3.63 $\pm$ 0.02       & 0.715 $\pm$ 0.004 \\
		Mean Teacher + UBPL                   & \pmb{3.65} $\pm$ 0.00 & \pmb{0.719} $\pm$ 0.001 \\
		\midrule
		DualPose                              & 3.90 $\pm$ 0.02       & 0.683 $\pm$ 0.002 \\
		DualPose + UBPL (noFDL)               & 3.81 $\pm$ 0.03       & 0.684 $\pm$ 0.001 \\
		DualPose + UBPL                       & \pmb{3.75} $\pm$ 0.07 & \pmb{0.686} $\pm$ 0.001 \\
		\bottomrule
		\toprule 
	\end{tabular}
	\caption{Results of the ablation experiment in pose estimation. Where, 'UBPL (noFDL)' means using our UBPL network and not using the Feature Decorrelation Loss (FD Loss).}
	\label{tab2_PoseAblationResults}
\end{table} 

The ablation results of our method on the semi-supervised pose estimation task are shown in Table. \ref{tab2_PoseAblationResults}. Without the use of Feature Decorrelation Loss (FD loss), UBPL can effectively improve the performance of semi-supervised methods. The results show that our approach can provide more accurate and informative pseudo-labels. When FD loss is used, the performance of the semi-supervised method is further improved. It shows that FD loss effectively improves diversity and makes the pseudo-labels generated by ensemble prediction more accurate.

\section{Experiments on Classification}

Further, we apply our method to semi-supervised classification tasks to validate the generality of our method.

\subsection{Implementation Details}

\begin{figure}[h]
	\centering
	\includegraphics[width=0.9\columnwidth]{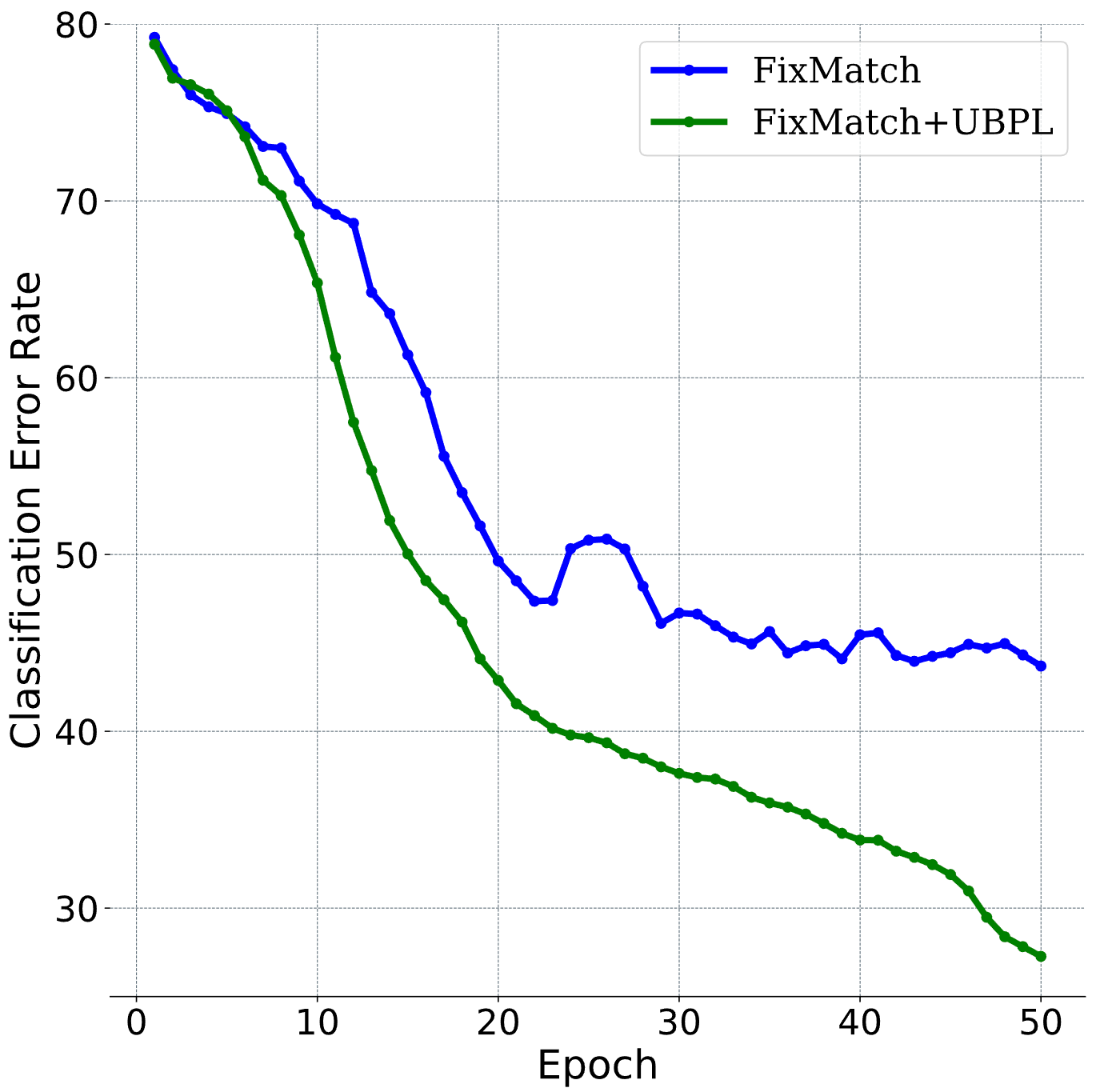}
	\caption{Error rate curve for 'FixMatch' and 'FixMatch+UBPL' on the CIFAR10 (40 labels).}
	\label{fig5_ClassResults}
\end{figure}

\begin{table*}[t]
\centering
\begin{tabular}{@{}lccccccccc@{}}
	\toprule 
	\multirow{2}{*}{Method} & \multicolumn{3}{c}{CIFAR10} & \multicolumn{3}{c}{CIFAR100} & \multicolumn{3}{c}{SVHN}\\
	& 40 & 250 & 4000 & 400 & 2500 & 10000  & 40 & 250 & 1000 \\
	\midrule
	FixMatch \cite{sohn2020fixmatch} & 43.69 & 12.21 & 10.28 & 81.15 & 46.42 & 32.20 & 89.71 & 79.94 & \pmb{77.86} \\
	FixMatch + UBPL & \pmb{27.27} & \pmb{11.44} & \pmb{9.29} & \pmb{80.88} & \pmb{44.96} & \pmb{31.37} & \pmb{90.00} & \pmb{79.00} & 78.36 \\
	\bottomrule
\end{tabular}
\caption{The error rate of semi-supervised classification on CIFAR10, CIFAR100 and SVHN datasets. 'Mean Teacher' and 'FixMatch' denote the use of them only, respectively. '+ UBPL' denotes using our UBPL network.}
\label{tab3_ClassMainResults}
\end{table*} 

We conduct experiments on the CIFAR-10/100 \cite{krizhevsky2009learning} dataset and SVHN dataset and extensively study the training performance under various labeled data amounts. 
In all experiments, the classification model is Wide ResNet (WRN) \cite{zagoruyko2016wide}. On the CIFAR-10 dataset, the widen-factor of WRN is 2. On the CIFAR-100 dataset, due to hardware limitations, the widen-factor of WRN is 6.

For a fair comparison, the same hyper-parameters are used for all experiments. Specifically, all experiments are performed with an standard Stochastic Gradient Descent (SGD) optimizer with momentum 0.9 \cite{sutskever2013importance, polyak1964some} and Nesterov momentum \cite{dozat2016incorporating} enabled.The learning rate is 0.03. The batch size of labeled data is 32, the batch size of labeled data is 224 (i.e., $ \mu=7 $). The confidence threshold $ \tau $ is 0.95. We used the same decay value (0.999) for the experiment involving the Exponential Moving Average (EMA) \cite{tarvainen2017mean}. Data augmentation for labeled samples consists of random horizontal flips and random cropping. Weak data augmentation for unlabeled samples includes random horizontal flipping and random clipping. Strong data augmentation adds RandAugment \cite{cubuk2020randaugment} to weak data augmentation, where n is 2 and m is 10.

$ \lambda_{ssl} $ is 10, $ \lambda_{pse} $ is 10. $ L_{fd} $ is 1000.

Due to hardware limitations, all experiments were trained for 50 epochs with 1024 iterations within each epoch. We used a fixed random seed (1388) to obtain accurate and reliable experimental results.

\subsection{Comparison with Existing Methods}

The classification error rate on the CIFAR-10/100 and SVHN dataset is shown in Table \ref{tab3_ClassMainResults}. The results show that our UBPL achieve better performance than FixMatch on most benchmark datasets. Our approach achieved better performance for tasks with extremely limited marker data (on CIFAR10 with 40 labeled data, on CIFAR100 with 400 labeled data).

As shown in Fig. \ref{fig5_ClassResults}, our UBPL performs better throughout the training period in the classification task, indicating that UBPL is towards unbiased and can generate better pseudo-labels.

\subsection{Ablation Studies}

\begin{table}[h]
	\centering
	\begin{tabular}{@{}lccc@{}}
		\toprule 
		\multirow{2}{*}{Method} & \multicolumn{3}{c}{CIFAR10} \\
		& 40 & 250 & 4000 \\
		\midrule
		FixMatch \cite{sohn2020fixmatch} & 43.69       & 12.21       & 10.28 \\
		\midrule
		FixMatch + UBPL (noFDL)          & 30.06       & 11.68       & 9.43 \\
		FixMatch + UBPL                  & \pmb{27.27} & \pmb{11.43} & \pmb{9.21} \\
		\bottomrule
		\toprule 
		\multirow{2}{*}{Method} & \multicolumn{3}{c}{CIFAR100} \\
		& 400 & 2500 & 10000 \\
		\midrule
		FixMatch                         & 81.15       & 46.42       & 32.20 \\
		\midrule
		FixMatch + UBPL (noFDL)          & 81.03       & 45.21       & 31.32 \\
		FixMatch + UBPL                  & \pmb{80.85} & \pmb{44.93} & \pmb{30.93} \\
		\bottomrule
	\end{tabular}
	\caption{Results of the ablation experiment in classification. Where, 'UBPL (noFDL)' means using our UBPL network and not using the Feature Decorrelation Loss (FD Loss).}
	\label{tab4_ClassAblationResults}
\end{table} 

The ablation results of the proposed method on the semi-supervised classification task are shown in Table. \ref{tab4_ClassAblationResults}. Without using FD loss, UBPL can improve the performance of semi-supervised methods to some extent. Moreover, the performance of the semi-supervised method is further improved when FD loss is introduced. It is shown that FD loss effectively improves the performance.

\section{Limitations}
Although our method performs better throughout the training period in both classification and regression tasks, it still suffers from shortcomings. Multi-model ensemble training is slow and consumes a lot of resources, which has certain limitations in practical applications. In the future, we will focus on how to improve training efficiency and reduce resource footprint.

\section{Conclusion}

In this paper, we first discuss the T2L Dilemma that arises when semi-supervised classification methods are directly applied to regression tasks. Then, we propose a theoretically guaranteed constraint for generating towards unbiased labels based on Chebyshev's inequality, combining multiple predictions to generate superior quality labels from several inferior ones. Specially, we propose an Unbiased Pseudo-labels network (UBPL network) with multiple branches to combine multiple predictions as pseudo-labels, where a Feature Decorrelation loss (FD loss) is proposed based on Chebyshev constraint. We conduct extensive experiments to validate the effectiveness of our method and show that it can be applied to both semi-supervised classification and regression. Our approach can be easily extended to any semi-supervised framework such as Mean Teacher, FixMatch, DualPose.

\bibliography{aaai24}

\end{document}